\documentclass[10pt, a4paper]{article}
\usepackage{authblk}
\usepackage{lrec-coling2024} 

\usepackage{arydshln}
\title{LongDocFACTScore: Evaluating the Factuality of Long Document Abstractive Summarisation}

\name{Jennifer A Bishop\textbf{$^{\mathrm{1}}$}, Qianqian Xie\textbf{$^{\mathrm{1}}$}, Sophia Ananiadou\textbf{$^{\mathrm{1,2}}$}} 

\address{\textbf{$^{\mathrm{1}}$}National Centre for Text Mining, Department of Computer Science, \\ The University of Manchester, United Kingdom\\
\textbf{$^{\mathrm{2}}$}Artificial Intelligence Research Center, Tokyo, Japan\\
         \{jabishop.research, xqq.sincere\}@gmail.com, sophia.ananiadou@manchester.ac.uk\\
         }

\abstract{
Maintaining factual consistency is a critical issue in abstractive text summarisation, however, it cannot be assessed by traditional automatic metrics used for evaluating text summarisation, such as ROUGE scoring. Recent efforts have been devoted to developing improved metrics for measuring factual consistency using pre-trained language models, but these metrics have restrictive token limits, and are therefore not suitable for evaluating long document text summarisation. Moreover, there is limited research and resources available for evaluating whether existing automatic evaluation metrics are fit for purpose when applied in long document settings. In this work, we evaluate the efficacy of automatic metrics for assessing the factual consistency of long document text summarisation. We create a human-annotated data set for evaluating automatic factuality metrics, LongSciVerify, which contains fine-grained factual consistency annotations for long document summaries from the scientific domain. We also propose a new evaluation framework, LongDocFACTScore, which is suitable for evaluating long document summarisation. This framework allows metrics to be efficiently extended to any length document and outperforms existing state-of-the-art metrics in its ability to correlate with human measures of factuality when used to evaluate long document summarisation data sets. We make our code and LongSciVerify data set publicly available: \url{https://github.com/jbshp/LongDocFACTScore}.
\\ \newline 
  }

\begin{document}

\maketitleabstract

\section{Introduction}

Factual inconsistency, i.e., when a generated summary is not entailed by its source document, is a well-documented limitation of modern neural summarisation methods \cite{maynez-etal-2020-faithfulness,wallace-2021-factual}. Although Large Language Models (LLMs) have shown greatly superior performance on a range of NLP tasks, including summarisation \cite{zhang2023benchmarking,xie2023survey}, and are increasingly being used for summarisation of long documents in real world applications, even the best performing models, such as GPT-4 \cite{openai2023gpt4}, are flawed in their ability to remain factual consistent \cite{bang2023multitask, ye2023flask, min2023factscore}.  

Human evaluation is generally regarded as the gold standard for evaluating generative models, yet it is timely and costly to conduct, particularly for tasks involving long documents, and thus only a small proportion of long document summarisation studies perform a human evaluation on long document data sets \cite{krishna-etal-2023-longeval}. Consequently, there is a requirement for effective automatic evaluation metrics which align to human judgement in long document settings. 

Although ROUGE scoring \cite{lin-2004-rouge} is the traditional metric for automatic evaluation of text summarisation, it is flawed and does not correlate well with human judgement \cite{yuan-2021,huang-etal-2020-achieved,kryscinski-etal-2019-neural}. There have been efforts to develop improved model-based metrics for measuring factual consistency \cite{scialom-etal-2021-questeval,yuan-2021,kryscinski-etal-2020-evaluating,qin2022t5score,fu2023gptscore, liu2023geval}, however, the studies proposing these metrics only conduct evaluation on short document summarisation data sets \cite{hermann2015teaching, grusky-etal-2018-newsroom,narayan-etal-2018-dont, pagnoni-etal-2021-understanding} and there is limited research or available data sets for evaluating these automatic metrics in long document settings. 

Since modern evaluation metrics use pre-trained language models (PLMs), they are only able to process a limited number of tokens at a time and must truncate, on average, over half of the tokens of a long source document in their calculations. Therefore, they cannot be applied effectively when used in long document settings \cite{koh22}. This issue is exacerbated when evaluating factual consistency, where many of the metrics are designed to be reference-free \cite{yuan-2021,fu2023gptscore, liu2023geval, scialom-etal-2021-questeval, kryscinski-etal-2020-evaluating}, i.e., they use the source document (generally a much longer document), rather than a gold summary, in their calculations.

In this work, we propose a reference-free evaluation framework, LongDocFACTScore, intended for assessing the factual consistency of abstractive summarisation of long documents. We show that this framework outperforms all other automatic metrics evaluated in their correlation with human annotations of factuality on the long document data sets in our experiments. 

Our proposed framework can be efficiently be extended to any length document and incorporates fine-grained, sentence-level assessments of factuality consistency to give a document-level score for the factual consistency of a summary. We conduct an evaluation of the efficacy and efficiency of LongDocFACTScore and other automatic evaluation metrics on a range of long and short document data sets.

Addressing the scarcity of resources for evaluating automatic metrics in long document settings, we create a long document data set of the scientific domain with fine-grained, expert, human annotations of factual consistency, which we make available alongside our code. We hope that this resource encourages future work into the evaluation of automatic metrics in long document settings.

\begin{figure*}
  \centering
      \includegraphics[width=0.8\linewidth]{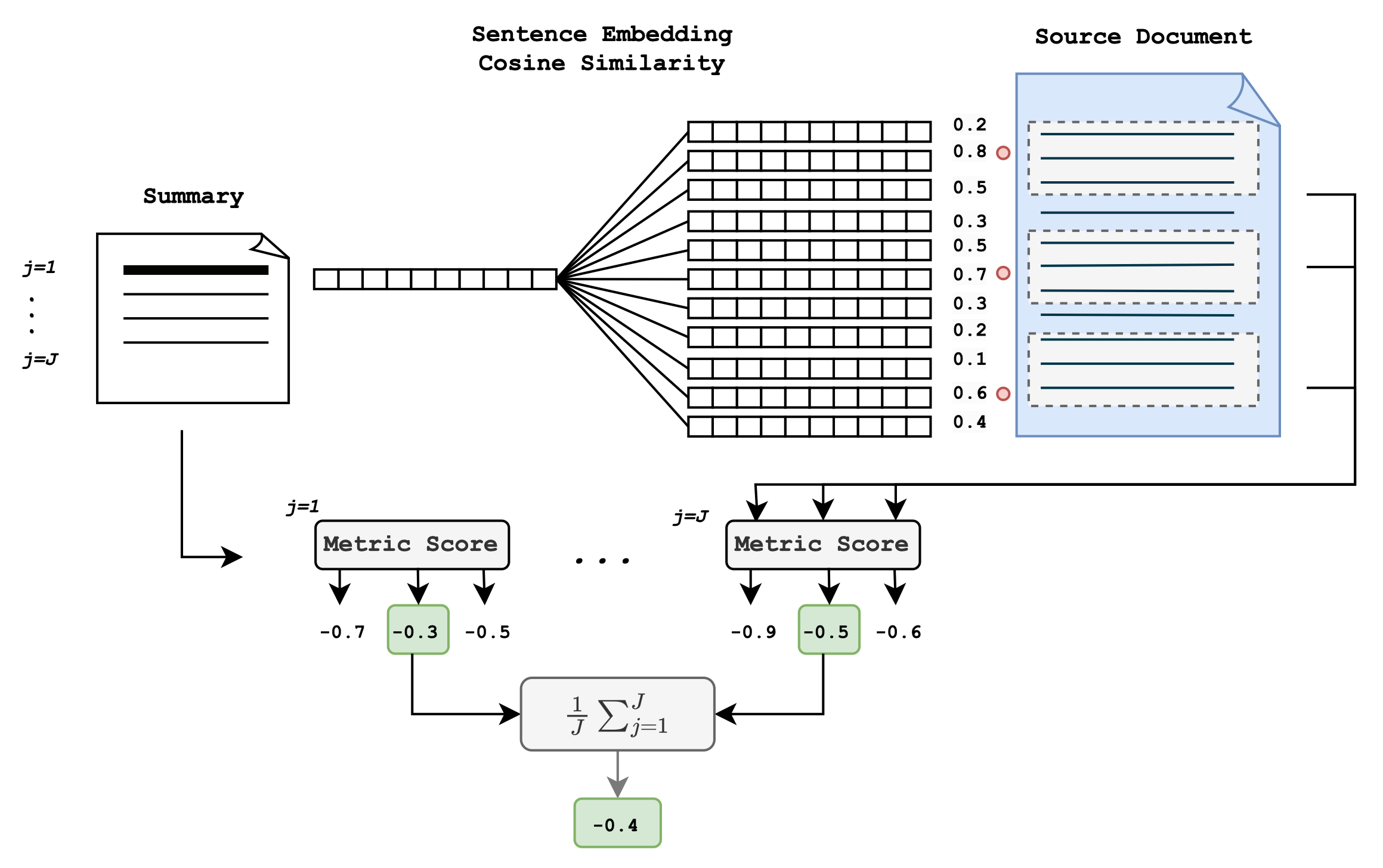}
       \caption{Illustration of the LongDocFACTScore framework.}
\label{fig:LongDocFACTScore}
\end{figure*}

\section{Related Work}

\subsection{Automatic Evaluation Metrics for Evaluating Factual Consistency}

ROUGE scoring \cite{lin-2004-rouge}, which uses word overlap between two texts to calculate their similarity, has long been the popular automatic metric used for evaluation of text summarisation. However, more recently, model-based metrics, such as BERTScore \cite{zhang2020bertscore}, which measures agreement at a token level between the cosine similarity of BERT-based \cite{devlin-etal-2019-bert} embeddings, have shown improved correlation with human judgement. Additionally, reference-free model-based metrics have shown improved performance for the evaluation of factual consistency on short document summarisation data sets. FactCC \cite{kryscinski-etal-2020-evaluating} uses a fine-tuned BERT-based classifier to predict, for each sentence of a summary, whether it is correct or incorrect, given its source document. QuestEval \cite{scialom-etal-2021-questeval} uses T5-based models \cite{raffel2020exploring} for a question generation and answering approach. BARTScore \cite{yuan-2021} uses BART \cite{lewis-etal-2020-bart} to calculate the log probability of generating a sequence of text, given a second sequence, to predict an automatic score. T5SCORE \cite{qin2022t5score} uses T5-based models and combines the generative approach taken by BARTScore with a discriminative approach - i.e., fine-tuning a model to predict a quality score. Unfortunately, many of these model-based metrics are costly to run, for example, QAGS \cite{wang-etal-2020-asking}, another question-answering based metric, running on a single NVIDIA v100 GPU, will take 4 days to process the CNN/DM test data set \cite{nan-etal-2021-improving}. Other recent works have proposed the use of LLMs for evaluation of NLP tasks \cite{fu2023gptscore,luo2023chatgpt,liu2023geval}, but still limit their evaluation to short document summarisation data sets. 

\subsection{Frameworks for Evaluation of Long Document Summarisation}

There has been limited research into automatic evaluation metrics for long document summarisation, despite the value of summarisation being derived mostly when applied to long documents. \citeauthor{koh22}, (\citeyear{koh22}) carried out a survey of long document summarisation and found a gap in research for automatic metrics which could efficiently and effectively be applied to long document data sets. \citeauthor{krishna-etal-2023-longeval}, (\citeyear{krishna-etal-2023-longeval}) propose guidelines for evaluation of long document summarisation, and provide LongEval, the only publicly available long document summarisation data set with human annotations of factual consistency which we could find at the time of conducting our research. SMART \cite{amplayo2022smart} proposes a method of extending evaluation metrics to long documents by cycling through them, but do not propose an efficient way of doing so, nor do they evaluate their work on long document summarisation data sets due to the lack of availability of these resources.

\section{Methods}

In this section, we describe the LongDocFACTScore framework. This framework builds on existing evaluation metrics but applies them in a novel way, providing a summary-level factuality score that considers fine-grained statements, whilst scaling efficiently though a document of any length. LongDocFACTScore evaluates each sentence in a predicted summary against the most similar sections of a source document, calculated using the cosine similarity of their sentence embeddings \cite{reimers-gurevych-2019-sentence}. Individual summary sentences are evaluated to ensure fine-grained statements are considered in the predicted summary-level score, whilst sentence embeddings are used to improve efficiency of the framework. 

To calculate LongDocFACTScore, both the source document  \( D  =\left\langle s_{i} , i \in I\right\rangle\) and its generated summary \( S =\left\langle s_{j} , j\in J\right\rangle\) are split into sentences using the nltk library\footnote{\url{https://www.nltk.org}}. Splitting a document into sentences before applying an evaluation metric has shown to be effective in prior works \cite{min2023factscore,amplayo2022smart}. For each of these sentences, sentence embeddings are generated using the sentence-transformers library\footnote{\url{ https://github.com/UKPLab/sentence-transformers}} initialised with the bert-base-nmli-mean-tokens model\footnote{\url{ https://huggingface.co/sentence-transformers/bert-base-nli-mean-tokens}}. For each sentence in the predicted summary \( s_{j}\), the cosine similarity between its sentence embedding and the sentence embedding of each sentence in the source document \( s_{i}\) is calculated. \( D\) is then re-indexed by the cosine similarity scores, so that the new index \( k\) is sorted by:

\begin{equation}
arg\max_{i\in I}\left(cosine\_similarity\left(s_{j}, s_{i}\right)\right).
\end{equation}

The \( K\) most similar source document sentences are then selected and are each concatenated with their preceding and following sentences, thus giving \( s_{k}^{\ast } = s_{k-1}+ s_{k}+ s_{k+1}\), to create the sequence of slightly longer text snippets. We select the \( K\) most similar source document text snippets to improve the efficiency of the framework. We assume the most similar source document text snippets are the ones most likely to be relevant to make an assessment of factual consistency.  

The metric score is then calculated between each of the source document text snippets \( s_{k}^{\ast }\) and the summary sentence \( s_{j}\), and the maximum of these scores is taken. In this work, we set  \(K = 3\), a decision which we justify in Section \ref{section:parameterstudy}.

For each sentence, \( s_{j}\) in \( S\), of the generated summary, the process is repeated, resulting in one score per generated summary sentence. The mean of these scores is then calculated, providing an overall summary score given by the equation:
\begin{equation}
 \frac{1}{J}\sum_{j = 1}^{J} \max_{k =\left\{ 1,2,3\right\} }(metric(s_{j}\left\vert s_{k}^{\ast }\right)).
\end{equation}

\begin{figure}
  \centering
      \includegraphics[width=1\linewidth]{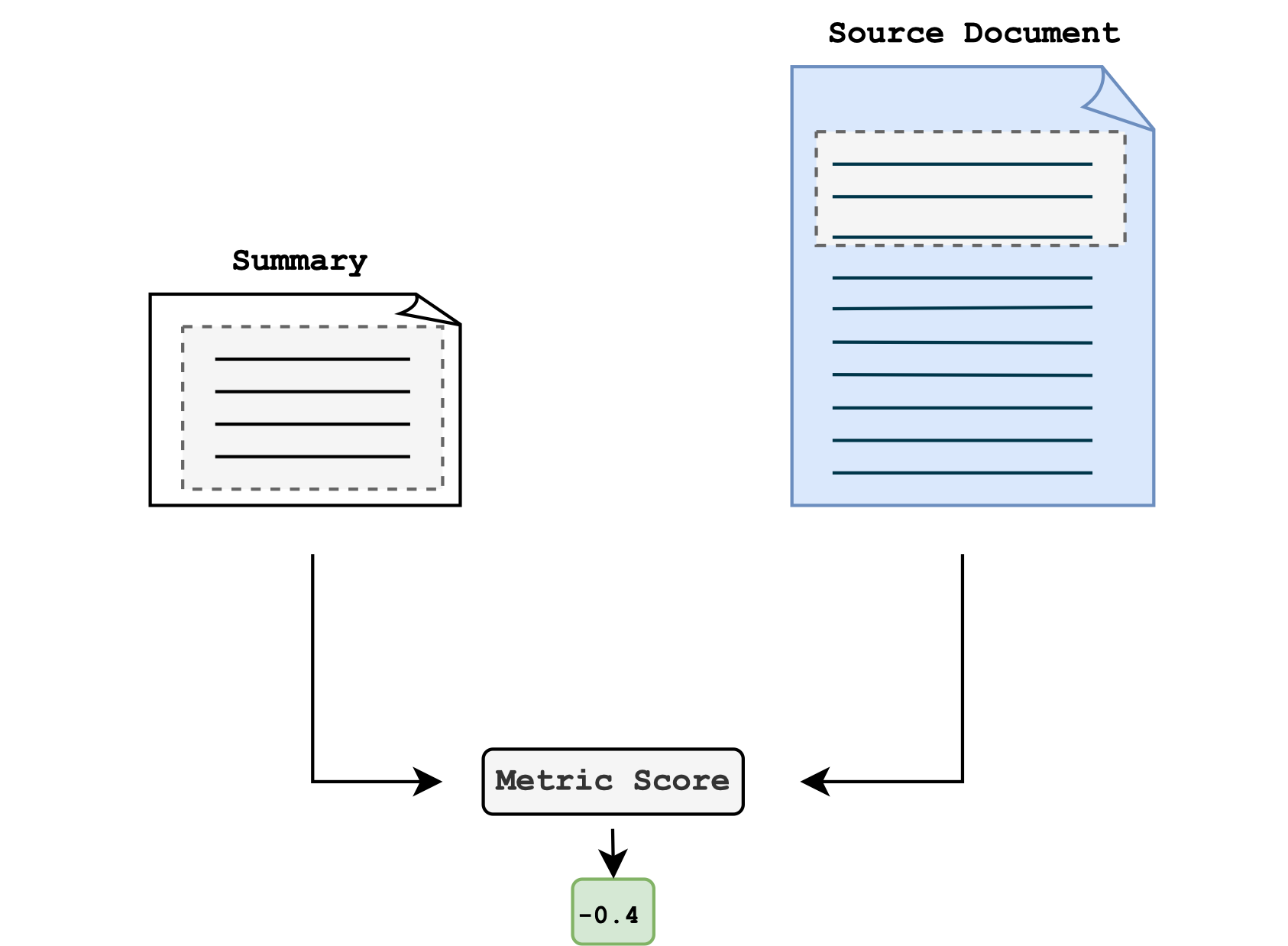}
       \caption{Calculation of a traditional automatic metric for assessing factual consistency.}
    \label{autom}
\end{figure}

Figure \ref{fig:LongDocFACTScore} illustrates this framework, showing for a single sentence in the generated summary, the similarity scores being calculated for every sentence in the source document, and the resulting three highest scoring sentences being concatenated with their surrounding sentences. A metric score is then calculated between these three source document text snippets and the summary sentence. As indicated in Figure \ref{fig:LongDocFACTScore}, this process is repeated for every sentence in the generated summary and the scores are averaged. For contrast, Figure \ref{autom} shows the method for directly applying an automatic scoring metric to a long document, without applying the LongDocFACTScore framework. The entire generated summary and the truncated long source document are directly input to the metric, resulting in one score. Consequently, there are three fundamental differences between LongDocFACTScore and an automatic metric designed for short document evaluation:
\begin{itemize}
    \item The first difference is that LongDocFACTScore considers sections of text from the full length of the source document in its calculation (using sentence embeddings to select the most relevant from across the document) whereas other metrics truncate the source document. For metrics applied without the LongDocFACTScore framework, if a generated summary includes content from the latter part of a long document, it will be ignored, which is a problem when assessing factual consistency of long document summarisation. 
    \item The second significant difference is that LongDocFACTScore calculates a metric score on short sections of text at one time, comparing one sentence in the predicted summary to a short section of the source document, rather than long, truncated sections. This allows for better evaluation of fine-grained statements. 
    \item Lastly, LongDocFACTScore uses sentence embeddings to identify the most similar parts of the source document. This improves the efficiency of the framework, as it avoids the metric needing to be applied for each pair-wise set of sentences in the source document and summary. 
\end{itemize}

\section{Experimental Data Sets}
\label{section:datasets}

We evaluate the automatic metrics on their ability to assess factual consistency on two long document data sets and several short document data sets. We collected our own long document data set, consisting of documents from the biomedical and scientific domains annotated by six expert human annotators with fine-grained factual consistency labels. We refer to this data set as the LongSciVerify data set and provide further details of its curation in Section \ref{section:longsciverify}. The data set is made available alongside our code. We further evaluate our methods on the LongEval PubMed data set \cite{krishna-etal-2023-longeval}, another long document data set with factual consistency annotations. Finally, we conduct an evaluation on a range of short document data sets with human annotations of factuality, which have been used to evaluate automatic metrics in prior works \cite{yuan-2021}.

\subsection{The LongSciVerify Data Set}
\label{section:longsciverify}

To support the evaluation of factuality metrics for long documents, we create a new data set called LongSciVerify, with multiple summaries generated from long documents, and fine-grained human annotation scores of their factual correctness. This data set consists of 270 annotated summaries generated from the long document, English-language PubMed and ArXiv data sets \cite{cohan-etal-2018-discourse}. A description of the PubMed and ArXiv data sets can be found in Table \ref{table:datasets}.

From each of the PubMed and ArXiv data sets, fifteen articles were randomly sampled. Summaries were generated for these data sets using three different abstractive methods, which were all able to consider the entire long document in the generation of their summaries. These methods were selected to enable an effective evaluation of the performance of the automatic metrics in long document settings. Details of the abstractive methods used to generate the summaries are provided in Appendix \ref{sec:appendix}. As the PubMed and ArXiv data sets included in this data set are highly domain specific, we recruited six expert annotators, three per data set, to review the automatically generated summaries. At the time of evaluation, all of the expert annotators reviewing the PubMed data set were, or were in the final years of study to be, qualified clinicians. The expert annotators for the ArXiv data set had all achieved a minimum of an undergraduate degree in a physical science. The annotators who participated in our study were colleagues of the authors and therefore volunteered to participate in the study without payment. It was made clear to the annotators that this human evaluation was for scientific research on abstractive summarisation with the intention for use in a scientific publication. 

\begin{table}[]
\centering
 \renewcommand{\arraystretch}{1.2}
\begin{tabular}{lcccc}
\hline
\textbf{} & \begin{tabular}[c]{@{}c@{}}Doc.\\ tokens\end{tabular} & \begin{tabular}[c]{@{}c@{}}Doc.\\ sentences\end{tabular} &\begin{tabular}[c]{@{}c@{}}Sum.\\ tokens\end{tabular}  & \begin{tabular}[c]{@{}c@{}}Sum.\\ sentences\end{tabular} \\
\hline
PM  & 3209  & 124  & 208   & 9 \\
AX  & 6515  & 249  & 279   & 11 \\ 
\hline
\end{tabular}
\caption{Average number of tokens and sentences in the evaluated data sets. PM denotes the PubMed data set and AX denotes the ArXiv data set.}
\label{table:datasets}
\end{table}

The definition of factual consistency we provided to annotators was taken from \citeauthor{fabbri-etal-2021-summeval}, (\citeyear{fabbri-etal-2021-summeval}): \textit{"Factual consistency: The factual alignment between the summary and the summarised source. A factually consistent summary contains only statements that are entailed by the source document."}. 

\begin{figure*}
  \centering
      \includegraphics[width=0.99\linewidth]{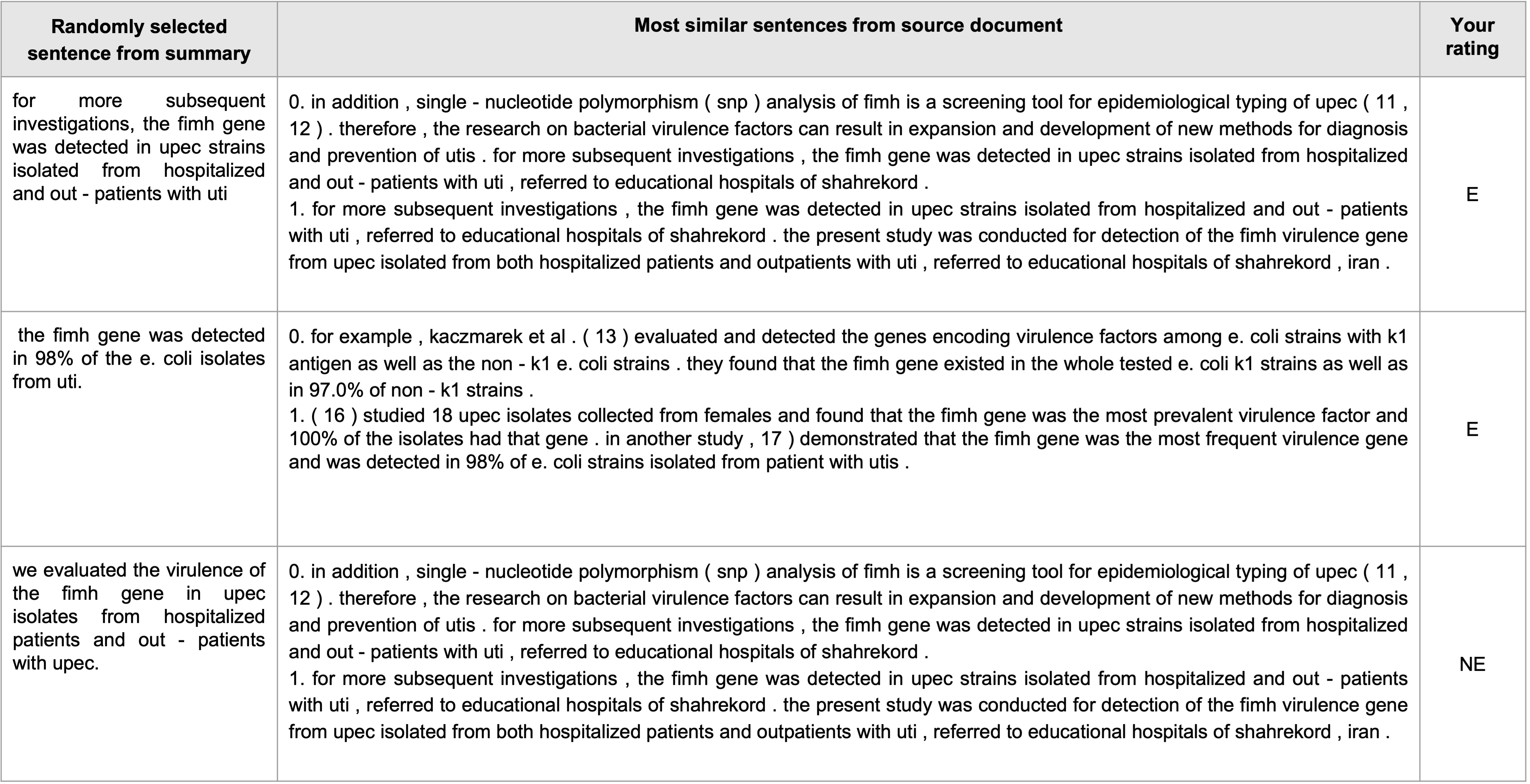}
       \caption{Example of a PubMed summary, annotated for factual consistency by an expert human annotator to create the LongSciVerify data set. "E" indicates that a sentence is "Entailed" and "NE" indicates a sentence is "Not Entailed".}
\label{fig:annotations}
\end{figure*}

We opted to capture a fine-grained binary classification metric (entailed vs not entailed), due to this having been shown to be effective and achieve higher inter-annotator agreement scores (IAA) in prior work \cite{krishna-etal-2023-longeval, min2023factscore}. Annotators were asked to mark a sentence as ‘not entailed’ if there were any factual inconsistencies. For each generated summary included in the study, we sampled three summary sentences and selected the most similar two text snippets (1-3 sentences) from the source document, calculated using sentence embeddings and cosine similarity. The human annotators were then given the three sentences sampled from the generated summary, and the corresponding two text snippets for each, from the source document and were asked to decide whether, given the text snippets, if each sentence was entailed or not. We provide an example screenshot of the factuality scoring for the three sampled sentences from a PubMed article summary in Figure \ref{fig:annotations}.

For each of the PubMed and ArXiv samples, each of the three human annotators evaluated the same three summaries (generated by the three different methods) from the same 15 randomly sampled documents, thus resulting in 270 annotated summaries, with three fine-grained annotations per summary. During the human evaluation study, the annotators were unaware of which method was used to create each summary. 

Table \ref{table:IAA} shows the IAA of the fine-grained human annotated data, for each data set, calculated using the Krippendorff’s alpha metric\footnote{\url{https://github.com/grrrr/krippendorff-alpha} } \cite{krippendorff2004}. In Table  \ref{table:IAA}, the IAA is calculated both between fine-grained, sentence-level entailment annotations, and for the summary-level annotations (i.e., the average of the fine-grained annotations per summary). For our LongSciVerify PubMed data set, the IAA of the fine-grained factual consistency annotations is relatively high. However, the IAA of the LongSciVerify ArXiv data set is a little lower. We hypothesise this could be due to the noise in the ArXiv data set \cite{koh22} and the highly domain-specific nature of the data set.

\begin{table}
\centering
 \renewcommand{\arraystretch}{1.2}
\begin{tabular}{cccc}
\hline
\textbf{} &
\begin{tabular}[c]{@{}c@{}}\textbf{LSV} \\ \textbf{ArXiv}\end{tabular} & \begin{tabular}[c]{@{}c@{}}\textbf{LSV} \\ \textbf{PubMed}\end{tabular}& \begin{tabular}[c]{@{}c@{}}\textbf{LE} \\ \textbf{PubMed}\end{tabular}\\
\hline
Fine-grained & 0.54 & 0.76 & - \\
Summary-level & 0.70 & 0.82 & 0.61 \\
\hline
\end{tabular}
\caption{IAA of the human-annotated data for the LongSciVerify (LSV) and LongEval (LE) data sets, calculated using Krippendorf's alpha metric, for fine-grained and summary-level annotations of factual consistency.}
\label{table:IAA}
\end{table}

\subsection{The LongEval Data Set}

We additionally evaluated LongDocFACTScore and the other automatic metrics included in our study on the publicly available long document PubMed LongEval data set \cite{krishna-etal-2023-longeval}. This data set consists of summaries generated from two abstractive models: LongT5-large \cite{guo-etal-2022-longt5} and BigBird-PEGASUS \cite{zaheer2020big}. Three expert annotators were hired to give annotations of factuality on 40 summaries (two summaries generated by different methods for 20 documents), giving 120 annotated summaries. The IAA of the summary-level human annotations of factual consistency is given in Table \ref{table:IAA}. As for the LongSciVerify data set that we create, the LongEval data set was created by averaging fine-grained annotations to give a summary-level factuality score. However, in constrast to the LongSciVerify data set, annotators considered the entire source article, rather than just the most relevant sections of it, when making their assessments.  

\section{Experiments}

\subsection{Experimental Setting}

As baselines, ROUGE\footnote{\url{https://huggingface.co/spaces/evaluate-metric/rouge}} and BERTScore were implemented. We additionally implemented SOTA reference-free metrics: FactCC\footnote{\url{https://github.com/salesforce/factCC}}, QuestEval\footnote{\url{https://github.com/ThomasScialom/QuestEval}}, and BARTScore\footnote{\url{https://github.com/neulab/BARTScore}} (using the ‘bart-large’ model\footnote{\url{https://huggingface.co/facebook/bart-large}}). In this work, we apply the LongDocFACTScore framework to extend the state-of-the-art (SOTA) metric BARTScore, also implemented with the ‘bart-large’ model. All experiments were run on a single NVIDIA v100 GPU and all metrics, apart from ROUGE, made use of the GPU compute.  For the long document data set evaluation, all metrics were applied in a reference-free setting, i.e., comparing the predicted summary to the source document. 

\subsection{Long Document Data Set Results}

To calculate the correlations between the human measure of factual consistency and automatic metrics, the fine-grained annotations were averaged to give a summary-level score. The summary-level human-annotated factuality scores were then averaged over the different annotators for each unique summary, thus giving a single summary-level human factuality score for each unique summary. These human-annotated, summary-level scores were then compared to summary-level scores predicted by each metric. Consequently, for each pair of metrics, the correlation is calculated between 45 summaries for each of the PubMed and ArXiv subsets of the LongSciVerify data set, and 40 summaries for the LongEval PubMed data set. 

\begin{table}[]
\centering
 \renewcommand{\arraystretch}{1.2}
\begin{tabular}{ccc}
\hline
\textbf{Metric} & \textbf{PubMed} & \textbf{ArXiv} \\
\hline
ROUGE-1      & 0.09   & 0.02           \\
ROUGE-2      & 0.29   & 0.17           \\
ROUGE-L      & 0.23   & 0.14           \\
BERTScore    & 0.24   & 0.27           \\
FactCC       & -0.06  & -0.08          \\
QuestEval    & 0.26   & 0.24           \\
BARTScore    & \underline{0.39}   & \underline{0.49}     \\
LongDocFACTScore      & \textbf{0.61}   & \textbf{0.61} \\
\hline
\end{tabular}
\caption{Kendall’s tau correlations between the human factual consistency annotations and automatic metrics for the LongSciVerify data set.}
\label{table:LSV-kt}
\end{table}

\begin{table}[]
\centering
 \renewcommand{\arraystretch}{1.2}
\begin{tabular}{cc}
\hline
\textbf{Metric} & \textbf{LongEval PubMed} \\
\hline
ROUGE-1      & 0.15                     \\
ROUGE-2      & \underline{0.26}                  \\
ROUGE-L      & 0.22                     \\
BERTScore    & 0.18                     \\
FactCC       & -0.14                    \\
QuestEval    & 0.13                     \\
BARTScore   & 0.22              \\
LongDocFACTScore      & \textbf{0.29}           \\
\hline
\end{tabular}
\caption{Kendall’s tau correlations between the human factual consistency annotations and automatic metrics for the LongEval PubMed data set.}
\label{table:LE-kt}
\end{table}

\begin{figure}
  \centering
      \includegraphics[width=0.99\linewidth]{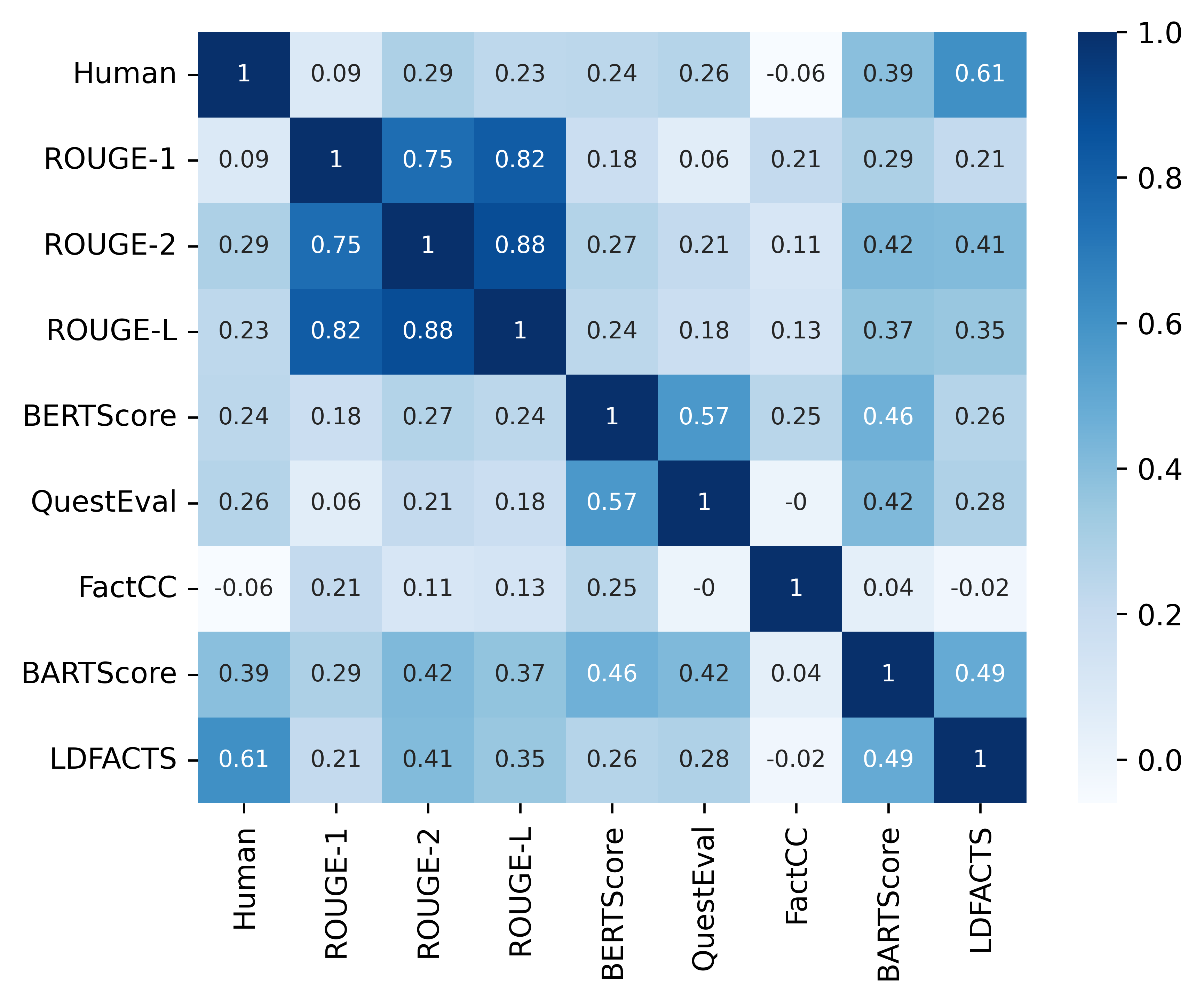}
       \caption{Pairwise Kendall’s tau correlations of metrics for the LongSciEval PubMed data set. LongDocFACTScore is denoted "LDFACTS", human annotations are denoted "Human".}
\label{fig:kt}
\end{figure}

Table \ref{table:LSV-kt} gives Kendall’s tau \cite{kendall1938new} correlations\footnote{\url{https://scipy.org}} between the human measures of factuality and the automatic metrics for the LongSciVerify data sets. Table \ref{table:LE-kt} gives the results of the same evaluation on the LongEval data set. Kendall’s tau correlations were calculated, rather than Spearman correlations, due to being more robust for data sets with smaller sample sizes. A pairwise correlation matrix between the automatic metrics and human annotations of factuality is given in Figure \ref{fig:kt}. We restrict this plot to only the LongSciVerify PubMed data set, due to this being the data set which achieves the highest IAA score in Table \ref{table:IAA}, and is therefore likely to be the most reliable data set for evaluation. In Table \ref{table:LSV-kt}, Table \ref{table:LE-kt}, and Figure \ref{fig:kt}, LongDocFACTScore, implemented by extending BARTScore, can be seen to correlate better with the human judgement of factual consistency than any other metric. Comparatively, we find that both FactCC and QuestEval show a low correlation with human judgement. BARTScore has a reasonable correlation with the human factual consistency annotations, however, since it is required to truncate the source document, we expect that it would become decreasingly correlated with human judgement as it is used to score texts of increasing length. ROUGE-2 and BERTScore perform best out of the baseline metrics evaluated, but no baseline metrics show a strong correlation with human measures of factual consistency. Interestingly, Figure \ref{fig:kt} shows that several automatic metrics have strong correlations with each other, suggesting that there is overlap in what they are measuring, but there is lower correlation between LongDocFACTScore and the other automatic metrics, suggesting that by providing coverage of a long document, LongDocFACTScore captures new information which the other metrics miss.

\subsection{Computational Efficiency}

In Table \ref{table:time}, we compare the average time taken, in seconds, to run each transformer-based \cite{vaswani2017attention} automatic metric designed to measure factual consistency on fifteen samples from the PubMed LongSciVerify data set. Table \ref{table:time} shows that LongDocFACTScore, implemented with BARTScore, is second fastest, despite evaluating the generated summary against the entire source document, rather than a truncated version of it. In contrast, QuestEval is shown to be 20x slower, and FactCC 3x slower, than LongDocFACTScore.

\begin{table}[]
\centering
 \renewcommand{\arraystretch}{1.2}
\begin{tabular}{cc}
\hline
\textbf{Metric} & \textbf{Time taken (s)} \\
\hline
FactCC       & 24                      \\
QuestEval    & 160                     \\
BARTScore    & \textbf{1}              \\
LongDocFACTScore      & \underline{8}  \\
\hline
\end{tabular}
\caption{Time (s) to run each metric on 15 samples.}
\label{table:time}
\end{table}

\subsection{Short Document Data Set Results}

 Although the intended use of LongDocFACTScore is to evaluate the factual consistency of abstractive summarisation for long documents, we additionally evaluate LongDocFACTScore against other automatic metrics on a variety of human annotated, short document, abstractive summarisation data sets, to validate its performance in this setting.  We repeat the analysis conducted by \citeauthor{yuan-2021}, (\citeyear{yuan-2021}) on the data sets containing human measures of factuality, and use their human annotated data and code\footnote{\url{https://github.com/neulab/BARTScore}}, to report the Spearman correlation results for the SummEval data set’s factuality measure \cite{fabbri-etal-2021-summeval}, the accuracy scores for the Rank19 data set’s factuality annotations \cite{falke-etal-2019-ranking}, and the Pearson correlation between the automatic metrics and the human factuality annotations for the two QAGS data sets \cite{wang-etal-2020-asking}. We used same the measures of correlation for each data set as in the original analysis conducted by \citeauthor{yuan-2021}, (\citeyear{yuan-2021}), rather than Kendall’s tau correlations, to enable a direct comparison to their reported scores. 
 
 Table \ref{table:shortresults} gives the results of this analysis. The top section of Table \ref{table:shortresults} gives the results reported by  \citeauthor{yuan-2021}, (\citeyear{yuan-2021}), in middle section we report our results, and in the bottom section we re-report  G-EVAL \cite{liu2023geval} results, a SOTA metric which uses GPT-4 to calculate factuality scores. In Table \ref{table:shortresults}, it can be seen that LongDocFACTScore performs comparably in its ability to measure factual consistency to the original BARTScore model for short document summaries, indicating that the framework can be used on documents of differing lengths. G-EVAL-4 reports higher correlations results on the SummEval and QAGS-XSUM data sets but slightly lower correlations on QAGS-CNN. They do not report results for the Rank19 data set. 
 
 Although GPT-4 has a much greater token limit than standard PLMs, there is still ultimately a limit, therefore, in future work, LongDocFACTScore could be used to extend this metric, or other LLM-based metrics to very long documents, or multi-document settings. Although the token limit is longer for LLMs, non-relevant content can distract LLM-based metrics and degrade performance \cite{min2023factscore}, therefore we hypothesise that applying the LongDocFACTScore framework, which considers smaller sections of text at one time, could improve the performance of these models. Future work which extends LLM-based metrics should also consider in its analysis the computational cost of running an LLM-based metric in comparison to running more efficient metrics. 
 
 \begin{table}[]
 \centering
 \small
 \renewcommand{\arraystretch}{1.2}
\begin{tabular}{ccccc}
\hline
  & \begin{tabular}[c]{@{}c@{}}SE\\ Fact\end{tabular} & \begin{tabular}[c]{@{}c@{}}R19\\ Acc\end{tabular} & \begin{tabular}[c]{@{}c@{}}QAGS\\ CNN\end{tabular} & \begin{tabular}[c]{@{}c@{}}QAGS\\ XSum\end{tabular} \\
\hline
ROUGE-1 & 0.16  & 0.57  & 0.34   & -0.01      \\
ROUGE-2 & 0.19  & 0.63  & 0.46   & 0.10 \\
ROUGE-L & 0.12  & 0.59  & 0.36   & 0.02       \\
MoverScore & 0.16  & \underline{0.71}  & 0.41   & 0.05       \\
BERTScore  & 0.11  & \underline{0.71}  & 0.58   & 0.02       \\
FactCC  & -     & 0.70  & -      & -          \\
QAGS    & -     & \textbf{0.72} & 0.55   & \underline{0.18}      \\
BARTScore  & 0.31    & 0.68    & \textbf{0.66}  & 0.01       \\
\hdashline
\begin{tabular}[c]{@{}c@{}}LDFACTS \\ (BARTScore)\end{tabular}  & \underline{0.36} & 0.68    & \underline{0.65}     & 0.04     \\
\hdashline
\begin{tabular}[c]{@{}c@{}}G-EVAL-4 \end{tabular} 
  & \textbf{0.51} & - & 0.63     & \textbf{0.56}     \\
\hline
\end{tabular}
\caption{Correlation between human measures of factuality on short document data sets, including re-reported results \cite{yuan-2021,liu2023geval}. LDFACTS denotes LongDocFACTScore.}
\label{table:shortresults}
\end{table}

\subsection{Parameter Study}
\label{section:parameterstudy}

We study effects that different parameter settings have on the LongDocFACTScore metric. We report the impact of different parameter settings on the Kendall’s tau correlations when evaluating the LongSciVerify data set. The PubMed and ArXiv articles are combined for this parameter study.  

Table \ref{table:params-K} shows the effect of varying \( K\) in the LongDocFACTScore framework (i.e., the maximum number of candidate similar source document sentences considered per summary sentence) on the Kendall's tau correlation with the human measures of factual consistency. The last row, \( K = I\), gives the Kendall's tau correlation when all sentences in the source document are considered. The correlation of BARTScore with human annotations of factual consistency is also provided for reference. \( K = 3\) is shown to be the best parameter, however, the effects of varying \( K\) are seen to be small. This is somewhat expected as the maximum score of the \( K\) text snippets is carried forward in the LongDocFACTScore metric, and it is likely that the highest scoring sentences correlate well with the most similar sentence embeddings. 

\begin{table}[]
\centering
 \renewcommand{\arraystretch}{1.2}
\begin{tabular}{cc}
\hline
\textbf{LongDocFACTScore setting} & \textbf{Score} \\
\hline
BARTScore                & 0.440          \\
LongDocFACTScore \( K = 1\)              & 0.605          \\
LongDocFACTScore \( K = 3\)             & \textbf{0.610} \\
LongDocFACTScore \( K = 5\)              & 0.600          \\
LongDocFACTScore \( K = 7\)              & 0.600          \\
LongDocFACTScore \( K = 9\)            & 0.595          \\
LongDocFACTScore \( K = 11\)           & 0.590         \\
LongDocFACTScore \( K = I\)              & 0.575         \\
\hline
\end{tabular}
\caption{The effect of varying \(K\), the number of similar sentences considered for the LongDocFACTScore calculation, on the Kendall’s tau correlation with human judgements of factuality.}
\label{table:params-K}
\end{table}

Although varying \( K \) is not shown make a large difference to the performance of the metric, by selecting \( K = 3\) candidate sentences, rather than cycling through all sentences in the source document (i.e., \( K = I\)), the score calculation in LongDocFACTScore is only calculated for approximately 1-2$\%$ of sentences from the source articles in the PubMed and ArXiv data sets. Therefore, by increasing the number of candidate similar sentences \( K\), LongDocFACTScore becomes decreasingly efficient and, by extension, less suitable for use on long documents.  To illustrate this point, in Table \ref{table:params-time} we give the results of the repeated efficiency calculation from Table \ref{table:time}, where LongDocFACTScore is implemented with \( K = 3\) and \( K = I\). If \( K = I\), there is no need to calculate sentence embeddings or perform the sentence similarity calculation, therefore we additionally report the time taken without these calculations. Table \ref{table:params-time} shows that, for the LongSciVerify PubMed long document data set, performing the sentence similarity calculation to select the \( K = 3\) most similar text snippets speeds up the metric over 15x. 

In Table \ref{ref:params-sents}, the number of candidate sentences is kept constant at \( K = 3\) and the effect of concatenating the source sentence with the previous and following sentence(s) to generate a text snippet is examined on the documents from the LongSciVerify data set. Table \ref{ref:params-sents} shows that although concatenating one sentence either side of a selected sentence performs best, there is little variation in the Kendall’s tau correlation between the different settings.

\begin{table}[]
\centering
 \renewcommand{\arraystretch}{1.2}
\begin{tabular}{cc}
\hline
\begin{tabular}[c]{@{}c@{}}\textbf{LongDocFACTScore} \\ \textbf{setting}\end{tabular}  & \textbf{Time taken (s)} \\
\hline
\( K = 3\)              & 8      \\
\( K = I\)             & 134 \\   
\begin{tabular}[c]{@{}c@{}}\( K = I\) \\ (no similarity calculations)\end{tabular}  & 125 \\ 
\hline
\end{tabular}
\caption{Time taken (s) to run LongDocFACTScore on 15 samples, when implemented with different settings.}
\label{table:params-time}
\end{table}

\begin{table}[]
\centering
 \renewcommand{\arraystretch}{1.2}
\begin{tabular}{cc}
\hline
\multicolumn{1}{c}{\textbf{Method}} & \textbf{Score} \\
\hline
\( s_{k}^{\ast } = s_{k}\)            & 0.605          \\
\( s_{k}^{\ast } = s_{k-1}+ s_{k}+ s_{k+1}\)                                    & \textbf{0.610} \\
   \( s_{k}^{\ast } = s_{k-2}+ s_{k}+ s_{k+2}\)                                 & 0.595   \\
\hline
\end{tabular}
\caption{The effect of varying the number of source document sentences concatenated for the LongDocFACTScore calculation on the Kendall’s tau correlation with human judgements of factuality.}
\label{ref:params-sents}
\end{table}

\section{Conclusion}

The prevalence of LLMs and other neural methods for abstractive summarisation of long documents in real world settings is rapidly increasing, however, the abstractive methods used to generate these summaries have known issues with factual inconsistency and hallucination. In this work, we begin to address the lack of research into the suitability of automatic evaluation metrics for assessing factual consistency of long document summarisation, and make the following contributions: (i) we show that existing automatic metrics for assessing factual consistency, which have previously shown good performance on short document data sets, do not perform well in long document settings, (ii) we propose a new framework, LongDocFACTScore, which is able to consider an entire source document in its calculation, without the need to truncate it, and outperforms existing SOTA metrics in its ability to correlate with human measures of factual consistency on long document summarisation data sets, whilst still being more efficient than many SOTA automatic evaluation metrics, (iii) we work to address the lack of resources for evaluating automatic metrics in long document settings and release our LongSciVerify data set, designed for evaluating factuality metrics on the long document summarisation task. We hope that this work promotes further research into automatic metrics for evaluating abstractive summarisation of long documents. In future work, we hope to apply the LongDocFACTScore framework to extend other automatic metrics, such as newer LLM-based metrics. We also hope to incorporate our work into wider LLM evaluation frameworks. 

\section*{Limitations}
Firstly, we review the limitations of our human evaluation study. In our study, we recruited expert annotators, as the long document data sets are domain specific. It is difficult to recruit large numbers of expert annotators and therefore an improvement on this work would be to conduct a larger human evaluation study with more annotators evaluating more documents.  We also note that two out of three annotators of the ArXiv data set have a first language which is not English, although they are both fluent in English. Furthermore, although the annotators of the ArXiv data set had all achieved a minimum of an undergraduate degree in a physical science, they did not necessarily study physics, which was the domain of most articles randomly sampled for human evaluation.

Secondly, we comment on the limitations of the LongDocFACTScore metric. One issue with this metric, and other SOTA factuality metrics, is that they favour extractive summaries. Therefore, although this metric is shown to be effective at measuring the factual consistency of long document abstractive summaries, we suggest that this metric is used in conjunction with other metrics, within a wider framework, such as FLASK \cite{ye2023flask}. We also note that this evaluation only included English-language data sets. 

Lastly, we discuss the computational cost of our work. We were able to monitor our GPU usage and found that for all experiments run in this period, we used approximately 1200 GPU hours. Despite our metric, LongDocFACTScore, being comparably efficient (see Table \ref{table:time} and Table \ref{table:params-time}), we acknowledge that working with large neural models, as well as having environmental implications, is not economically possible for many researchers.

\section*{Ethics Statement}
Throughout our research, we complied with our institution’s ethical guidelines. We used open-source data and software, for which no ethical approvals were required. 

In our study, we conduct a human evaluation. As detailed in Section \ref{section:datasets}, we were fortunate enough to be able to recruit colleagues, who are domain-experts in the field of the data sets. They volunteered to participate in the study without payment, so we did not need to consider the ethics of crowd-worker payment. 

Our work proposes a metric for assessing the factual consistency of abstractive summaries generated for long documents. This metric can be used to help researchers assess the performance of their summarisation methods, however, to minimize any harm which may be caused by deploying an abstractive summarisation model in a live setting, we suggest that any method should be thoroughly evaluated by humans in the setting it is intended to be deployed.

\section*{Acknowledgements}
This work was partially supported by the project JPNP20006 from New Energy and Industrial Technology Development Organization (NEDO).

\section{Bibliographical References}\label{sec:reference}

\bibliographystyle{lrec-coling2024-natbib}
\bibliography{anthology,custom}

\clearpage
\newpage

\appendix

\section{LongSciVerify: Abstractive Summarisation Methods}
\label{sec:appendix}

For our human evaluation, we provide summaries generated using three different abstractive summarisation methods, which all consider text from across the entire length of a long document when generating a summary. 

\begin{figure}
  \centering
      \includegraphics[width=0.99\linewidth]{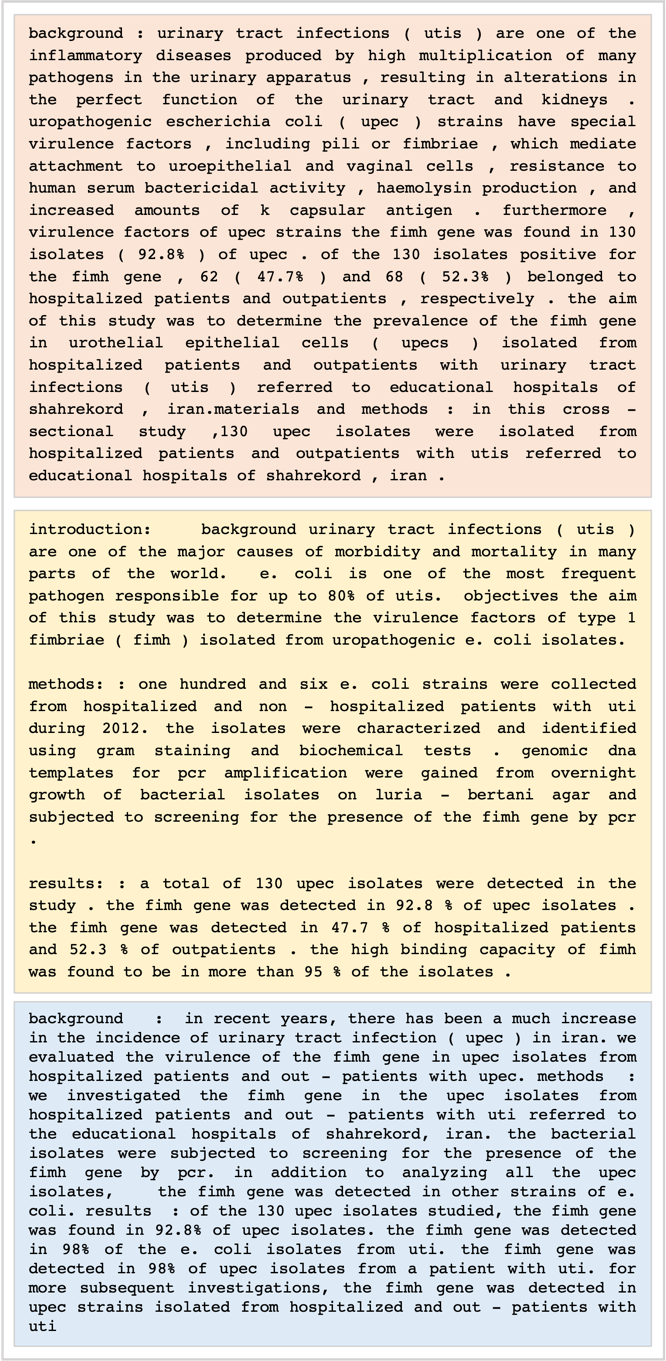}
       \caption{Automatically generated summaries included in our human evaluation study for one article sampled from the PubMed data set.}
\label{sum_method}
\end{figure}

Text zoning is a task which aims to segment a larger body of text into different zones or sections \cite{teufel1999argumentative}. Two of the methods we implement apply text zoning, and treat the identified sections independently, so that a PLM used to train the abstractive summarisation model is only required to process a document section at a time, rather than the entire document, to avoid truncation. We first implement DANCER \cite{gidiotis-2020}, a SOTA method which fine-tunes the PEGASUS PLM \cite{zhang2020pegasus}. DANCER splits a document into zones using keyword matching, then finds corresponding sections of the target abstract using ROUGE matching \cite{lin-2004-rouge}. It then uses beam search decoding to generate the summaries and combines the generated summaries of each section to form the article summary. An example output of the DANCER method can be seen in the top block of Figure \ref{sum_method}. The second method we implement is a method we develop. It applies a similar zoning approach to DANCER, but with two notable differences. Firstly, the sections of the summary used to create the target training pairs are matched by using both keywords and assumptions about the structure of the long scientific documents used in our study \cite{cohan-etal-2018-discourse}. Consequently, the target sentences for each section are always sentences which were consecutive to one another in the original summary, thus resulting in a summary which follows the logical structure of the document. Secondly, the summaries generated do not use beam decoding and are highly structured as each section of the summary is prefixed with the type of section it was generated from, e.g., ‘\textit{results:}’. An example of a summary generated by this method can be seen in the middle block of Figure \ref{sum_method}. The last method we implement uses an extractive-abstractive approach. We train our extractive-abstractive model using ORACLE extractive summaries as an input, optimized for a recall metric but limiting the number of sentences selected so that the total number of input tokens is less than 1024. At test time, we implement the unsupervised, extractive method GenCompareSum \cite{bishop-etal-2022-gencomparesum}, which does not truncate the source document and has previously shown strong performance on the PubMed and ArXiv data sets. An example of a summary generated with this method can be seen in the final block of Figure \ref{sum_method}. We use the train/val/test splits from the original data sets and use DANCER and GenCompareSum with their default settings. For our zoning method and the extractive-abstractive method, we fine-tune the LED PLM \cite{beltagy2020longformer} for three epochs with its default parameters. All experiments are run on a single NVIDIA v100 GPU.

\end{document}